\documentclass[pdflatex,iicol,sn-mathphys-num]{sn-jnl}

\usepackage{graphicx}
\usepackage{multirow}
\usepackage{amsmath,amssymb,amsfonts}
\usepackage{booktabs}
\usepackage{xcolor}
\usepackage{url}
\usepackage{xurl}
\usepackage{ragged2e}
\usepackage{etoolbox}

\AtBeginDocument{%
  \setlength{\topmargin}{-0.70in}%
  \setlength{\headheight}{10pt}%
  \setlength{\headsep}{6pt}%
  \setlength{\textheight}{10.35in}%
  \setlength{\footskip}{14pt}%
}
\begin{document}

\title[A Controlled Study of CLIP-Based Body-Scene Fusion]{A Controlled Study of CLIP-Based Body-Scene Fusion for Emotion Recognition in Context}

\author*[1]{\fnm{Zubair} \sur{Abbas} {\small\href{https://orcid.org/0009-0002-3906-0580}{ORCID}}}\email{l1f22bscs0304@ucp.edu.pk}

\author[1]{\fnm{Muhammad} \sur{Umair} {\small\href{https://orcid.org/0000-0002-6061-099X}{ORCID}}}\email{dr.umair@ucp.edu.pk}

\author[1]{\fnm{Muqaddas} \sur{Hameed} {\small\href{https://orcid.org/0009-0007-0875-3460}{ORCID}}}\email{l1s22bscs0383@ucp.edu.pk}

\affil*[1]{\orgdiv{Department of Computer Science}, \orgname{University of Central Punjab}, \orgaddress{\city{Lahore}, \country{Pakistan}}}

\abstract{Apparent emotion in natural images is often not visible from the face alone. The face may be small, hidden, or neutral, while posture and scene context carry much of the evidence. This work studies context-aware emotion recognition on EMOTIC with an image-only two-stream model. A ResNet-18 body stream encodes the target-person crop, and a CLIP ViT-B/16 scene stream encodes the full image. The fused feature predicts 26 categorical emotion labels and the continuous valence, arousal, and dominance values.

This study examines whether small context-debiasing or rare-class training changes still help after adding a CLIP scene encoder. The clean two-stream model is compared with simplified CCIM-style intervention, CLEF-lite context-bias subtraction, ASL tuning, and class-balanced sampling under the same implementation pipeline. No tested variant improves over the clean two-stream model, which achieves 34.52\% mAP on the EMOTIC test split. CLIP gives the model broad scene semantics, but the simplified causal, counterfactual, and rare-class changes do not automatically improve performance. Most remaining errors are in rare and subtle emotion categories, so the next step should focus on label relationships and finer subject-context interaction.}

\keywords{Emotion recognition, affective computing, EMOTIC, context-aware recognition, CLIP, multi-label classification}

\maketitle

\section{Introduction}
Emotion recognition in the wild is messier than facial expression recognition in a controlled image. A person may be turned away from the camera, partly occluded, far from the lens, or wearing a neutral face while the surrounding scene tells a different story. A runner at the finish line, a child at a birthday party, and a patient in a hospital room are not understood only from pixels inside the face. Context matters. That is why emotion recognition in context is not just a harder face-recognition problem.

EMOTIC was built for this setting. It contains people in natural images, annotated with 26 emotion categories and three continuous affect dimensions: valence, arousal, and dominance \cite{kosti2017emotion,kosti2020context}. The original EMOTIC work already showed that both the target person and the surrounding scene help. Later work pushed the context side further through multimodal cues, semantic scene information, object relations, causal debiasing, counterfactual inference, depth, and single-stage subject-context transformers \cite{mittal2020emoticon,wang2023scene,yang2023contextdeconfounded,yang2024clef,li2024dsct}.

Those directions are valuable, but they also make the pipeline heavier. Recent methods add semantic scene information, depth-based social interaction, causal debiasing, counterfactual inference, or transformer-based subject-context interaction. CLIP-style emotion representation learning has also been explored through larger pretraining frameworks \cite{zhang2023emotionclip,chen2024uniemox}. This work asks a narrower version of the problem. If a directly supervised image-only model already uses an off-the-shelf CLIP visual encoder for scene context, do extra simplified context-debiasing or rare-class training changes still help?

CLIP is attractive here because it gives a direct way to encode broad scene meaning. A separate semantic, object, or scene-context stream is not required before the classifier sees the image. Prior methods such as EmotiCon improve EMOTIC performance by combining multiple context sources, including personal cues, semantic scene information, and socio-dynamic cues. That richer design is valuable, but it is also more complex than a two-stream image model. The central comparison is whether a lighter design can capture part of the same contextual information by using CLIP as the full-scene encoder. This does not mean that CLIP replaces depth, object relations, or person-person interaction. It does not. The study measures how far a CLIP-based scene branch can go as an alternative to some of those heavier context streams.

Based on this motivation, the study is organized around three research questions:
\begin{itemize}
    \item \textbf{RQ1:} How effective is a lightweight two-stream body-scene model when CLIP ViT-B/16 is used as the full-scene context encoder?
    \item \textbf{RQ2:} Under the same implementation pipeline, do simplified CCIM-style intervention, CLEF-lite context-bias subtraction, ASL tuning, or class-balanced sampling improve over the clean CLIP-based two-stream baseline?
    \item \textbf{RQ3:} Which emotion categories remain difficult after CLIP-based body-scene fusion, and what do these class-level results imply for future lightweight improvements?
\end{itemize}

A controlled two-stream model is built for this purpose. The body stream uses ResNet-18 on the target-person crop. The context stream uses CLIP ViT-B/16 on the full image. Around this baseline, the experiments test simplified CCIM-style intervention, CLEF-lite output subtraction, ASL tuning, and class-balanced sampling. The clean two-stream model gives the highest mAP among the tested variants, with 34.52\% mAP on the EMOTIC test split.

The main contribution of this work is a controlled CLIP-based empirical study, not a claim that two-stream fusion is new. The contributions are:
\begin{itemize}
    \item A lightweight image-only EMOTIC baseline is built by combining target-person body features with CLIP-based full-scene context.
    \item The clean baseline is compared with simplified CCIM-style, CLEF-lite, ASL tuning, and class-balanced sampling variants under the same pipeline.
    \item The experiments show that no tested extension improves over the clean CLIP-based two-stream model, so simplified add-ons do not necessarily improve this CLIP-based scene baseline.
    \item Class-level AP is analyzed to identify rare and subtle categories as the main weakness, motivating label-correlation and fine-grained subject-context modeling.
\end{itemize}

\section{Related Work}
Context-aware emotion recognition has grown through several overlapping lines of work. Some papers treat the person and the scene as two complementary visual inputs. Others add pose, skeletons, depth, or semantic scene structure. More recent work tries to reduce context bias or learn subject-context interaction more directly. The CLIP-based two-stream study in this paper sits between the simple body-scene models and the heavier context-modeling systems.

\subsection{Emotion Recognition in Context}
Kosti et al. introduced EMOTIC and showed that emotion recognition improves when the model sees both the target person and the scene \cite{kosti2017emotion,kosti2020context}. That shift matters because the task is no longer only about facial expression. It is about a person inside a situation. CAER-Net follows a related idea by combining face information with hidden-face context, so surrounding visual cues can carry the prediction when facial expression is weak or unavailable \cite{lee2019caernet}.

Body and context cues have also been studied outside the exact EMOTIC setup. Huang et al. used a three-branch body, skeleton, and context model for emotion recognition in the wild \cite{huang2021bodycontext}. For this work, the exact branch design is less important than the assumption behind it: posture and surrounding scene information can remain informative when the face is blurred, small, or missing.

EmotiCon is a more complex reference point. It combines personal cues, semantic scene information, and socio-dynamic cues such as depth and interaction between agents \cite{mittal2020emoticon}. Its reported 35.48 AP on EMOTIC is close enough to the result in this paper to be worth discussing, but the architecture is not a clean match for the image-only two-stream setup used here. It is treated as evidence that richer context streams can help, not as a model reproduced in this work.

\subsection{Scene Semantics and Vision-Language Models}
Scene semantics are another route into the same problem. Wang and Sankaranarayana combined scene and semantic features with personal features on EMOTIC, using objects, attributes, and relationships as emotional evidence \cite{wang2023scene}. The model in this paper is simpler than that. CLIP can encode broad scene semantics, but it does not explicitly tell the classifier which object the person is using, which person they are facing, or which relation matters.

Large vision-language models take the semantic idea further. Etesam et al. used LVLMs and language reasoning over the target person in a bounding box \cite{etesam2024contextual}. Lei et al. explored LVLMs for context-aware emotion recognition through fine-tuning, few-shot prompting, and reasoning-style inference \cite{lei2024lvlm}. These papers show the value of language-aligned visual representations for emotion-in-context, but they sit in a different cost range. The CLIP use in this work is lighter: the visual encoder becomes the scene branch inside a trainable two-stream model.

CLIP-based emotion representation learning is close to the motivation of this work, but not identical. EmotionCLIP learns emotion-aware visual representations from verbal and nonverbal communication data, while UniEmoX studies cross-modal emotion representation learning with CLIP-style semantic knowledge \cite{zhang2023emotionclip,chen2024uniemox}. Those works use broader pretraining objectives or auxiliary data. This study does not add a new emotion-pretraining stage. The CLIP visual encoder is taken off the shelf, used as the full-scene branch, and fine-tuned inside the supervised EMOTIC model.

In this sense, CLIP replaces part of the semantic-context machinery, not every context module. EmotiCon explicitly separates personal, semantic, and socio-dynamic context. Other methods use object attributes, relationships, or scene parsing. The present model keeps only the body crop and the full image, then lets CLIP carry as much high-level scene information as it can.

\subsection{Context Bias and Causal Debiasing}
Context can help and hurt at the same time. A model may overlearn that a hospital means sadness or that a playground means happiness, even when the target person does not match that shortcut. CCIM treats this context bias as a confounder and applies a causal intervention module \cite{yang2023contextdeconfounded}. CLEF attacks a related issue by removing the direct context effect through factual and counterfactual comparison \cite{yang2024clef}. Both methods report consistent gains when the full framework is implemented carefully.

Those papers shaped the ablations in this study. The simplified CCIM-style and CLEF-lite versions, however, did not improve the CLIP two-stream baseline. This is not a claim against CCIM or CLEF themselves. It is a narrower finding about lightweight adaptations, where the exact branch design, training protocol, and counterfactual signal are not the same as in the original papers.

\subsection{Structured Context and Relation Modeling}
Several recent papers move past global scene features and model relationships inside the image. Prior work has used semantic scene cues, object relationships, depth-based interaction cues, and agent-scene interaction \cite{wang2023scene,mittal2020emoticon,li2024dsct}. The shared idea is straightforward: emotion is not only tied to the background category, but also to how the target person relates to objects, other people, and the ongoing activity.

This matters because the fusion block in this work uses global body and scene features. It does not know which object the person is holding, whether another person is nearby, or which region of the scene is emotionally relevant. Relation modeling may be a better next step than simply adding another global context stream.

\subsection{Fine-Grained Subject-Context Interaction}
Another route is to make subject-context interaction part of the model itself. DSCT replaces the usual two-stage pipeline with a single-stage framework that jointly handles subject localization and emotion classification through decoupled subject and context queries \cite{li2024dsct}. That design can learn fine-grained subject-context relations instead of waiting until the end to fuse two global vectors. It is also a much larger change than this project.

\subsection{Research Gap}
An open uncertainty is how much extra context modeling is still needed when the scene branch is already CLIP-based. Context-aware emotion recognition has benefited from body posture, scene semantics, object relations, depth, causal debiasing, and transformer-based subject-context interaction. Many of those gains come from larger pipelines, extra supervision, or carefully designed causal and counterfactual training.

This makes it hard to separate the value of the context idea from the complexity of the full method. It is also unclear whether simplified debiasing modules or rare-class training changes still help in a lightweight image-only setting. This matters for EMOTIC because many difficult labels are subjective, visually ambiguous, or underrepresented.

\section{Methodology}
All experiments use the same basic pipeline. For each annotated subject in EMOTIC, two visual inputs are prepared: a crop of the target person for the body stream and the full image for the scene stream. The body crop is processed by ResNet-18, while the full image is processed by CLIP ViT-B/16. The two features are fused and used for both multi-label emotion classification and VAD regression. The rest of this section describes each part of that pipeline, starting from the dataset and preprocessing, then moving to the model, loss function, ablations, training setup, and evaluation metrics.

\subsection{Dataset}
The experiments use EMOTIC. Each annotated person has a bounding box, one or more categorical emotion labels, and valence-arousal-dominance scores. The categorical problem is multi-label, so a single subject can be annotated as both Engagement and Happiness, or with other co-occurring categories. The continuous VAD values are normalized to $[0,1]$.

\begin{figure}[!htbp]
\centering
\includegraphics[width=0.96\linewidth]{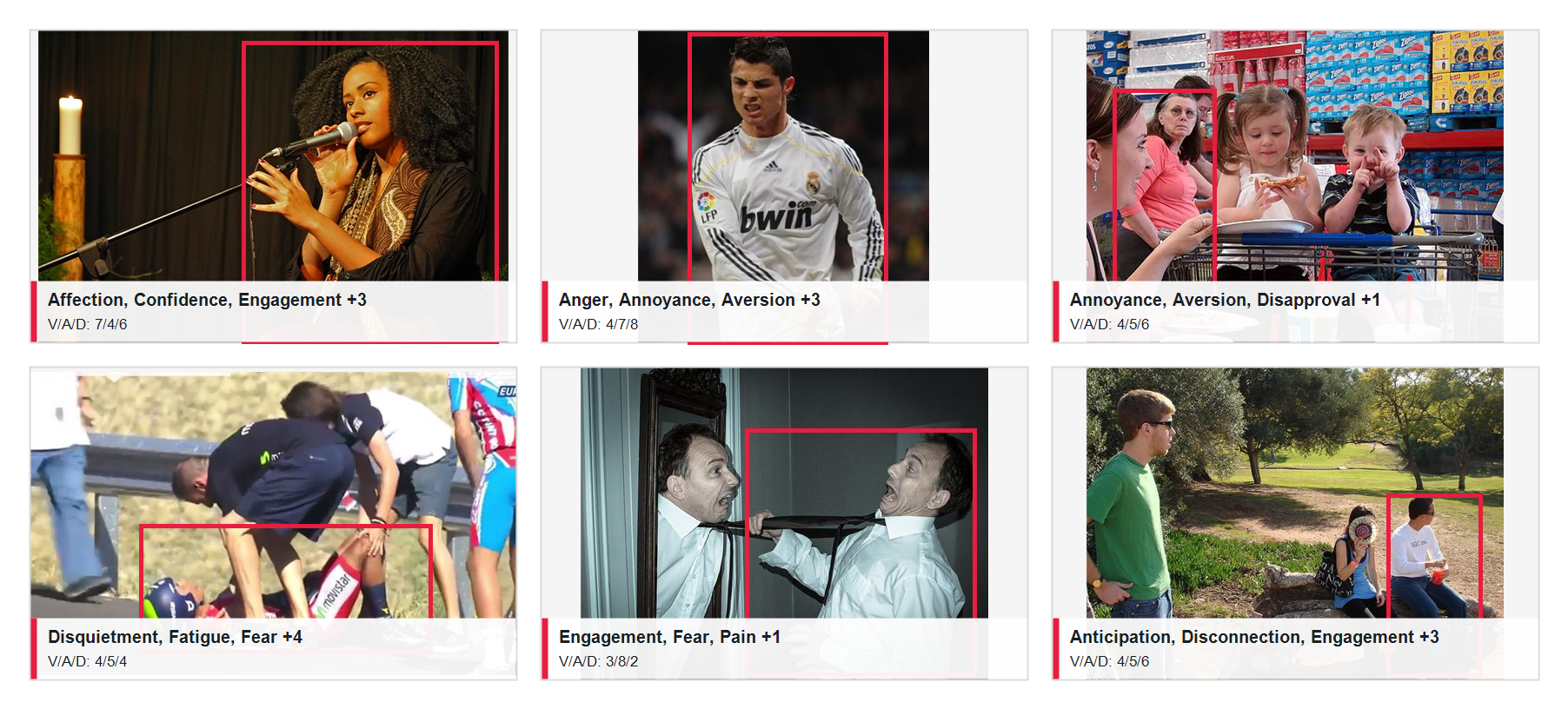}
\caption{Example EMOTIC samples with target-person bounding boxes, categorical emotion labels, and VAD annotations.}
\label{fig:dataset}
\end{figure}

Table~\ref{tab:splits} gives the train, validation, and test split sizes used in the implementation. Fig.~\ref{fig:dataset} shows EMOTIC examples with target-person boxes and annotations.

\begin{table}[!htbp]
\centering
\caption{EMOTIC split sizes used in the implementation.}
\label{tab:splits}
\begin{tabular}{lr}
\toprule
Split & Annotated subjects \\
\midrule
Train & 23,706 \\
Validation & 3,334 \\
Test & 7,280 \\
\bottomrule
\end{tabular}
\end{table}

\subsection{Preprocessing}
For each annotated person, two image inputs are created. The target-person crop is resized to $128\times128$ and normalized with ImageNet statistics \cite{deng2009imagenet}. The full image is resized to $224\times224$ and normalized with CLIP statistics \cite{radford2021clip}. During training, the body crop receives stronger augmentation because clothing, pose, scale, and crop quality vary widely. The context image receives milder augmentation so that the scene semantics remain recognizable.

\begin{figure}[!htbp]
\centering
\includegraphics[width=0.92\linewidth]{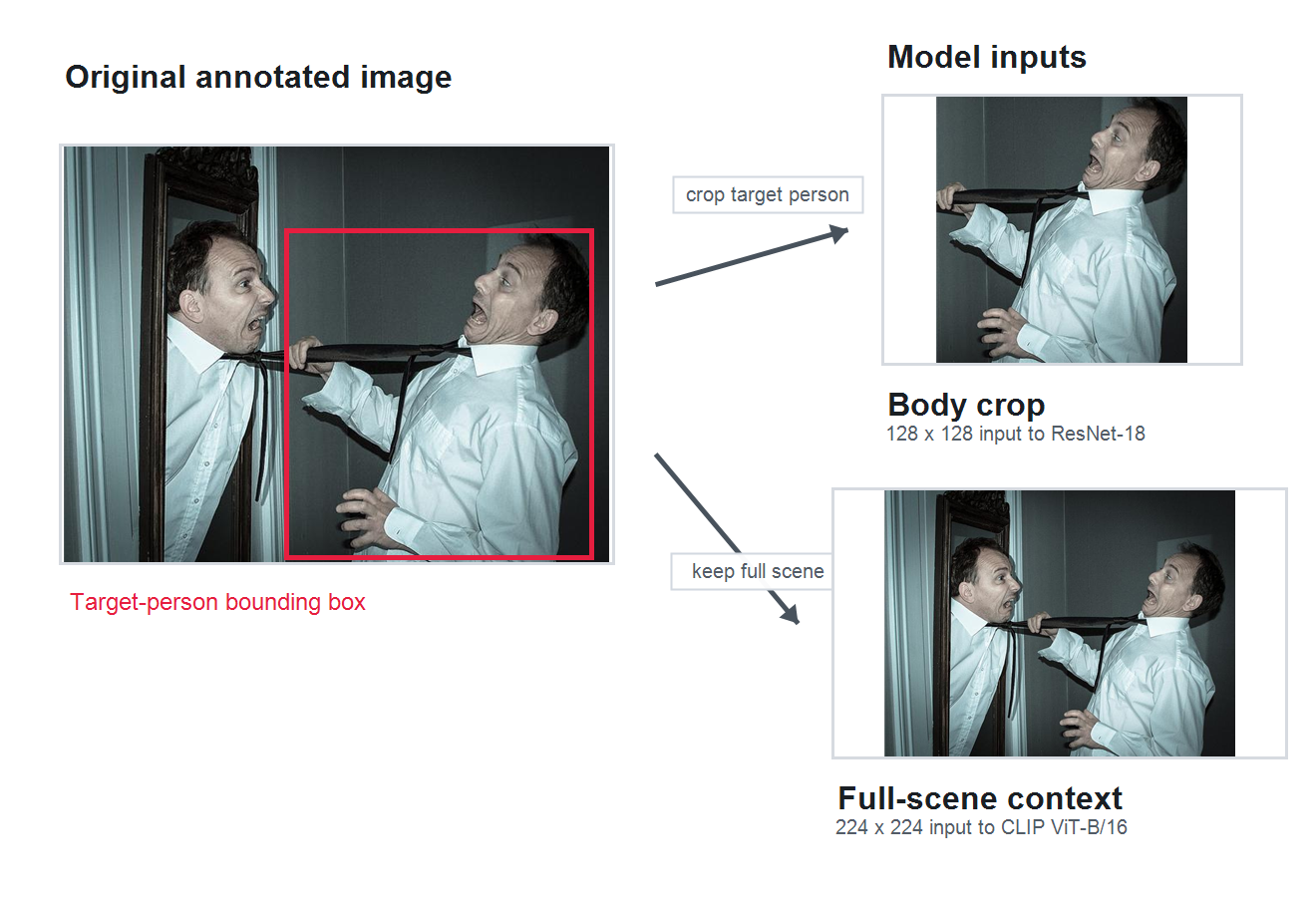}
\caption{Preprocessing pipeline used to form the two model inputs: a target-person body crop for the ResNet-18 stream and the full image for the CLIP scene stream.}
\label{fig:preprocess}
\end{figure}

Fig.~\ref{fig:preprocess} summarizes the preprocessing flow. The same original image is used twice: once as a focused body crop and once as a full-scene context view.

\subsection{Two-Stream Architecture}
The final model has two streams. The body stream is an ImageNet-pretrained ResNet-18 \cite{he2016resnet,deng2009imagenet}; its final classification layer is removed, and the extracted feature is projected to 256 dimensions. The context stream is CLIP ViT-B/16 \cite{radford2021clip,vaswani2017attention}, also projected to 256 dimensions. During fine-tuning, the top four CLIP transformer blocks are unfrozen.

After projection, the two streams are fused at the feature level. The fused representation feeds two heads: one head produces logits for the 26 categorical emotions, and the other predicts VAD.

Formally, for each target person, the body crop is denoted as $x_b$ and the full-scene image as $x_s$. The body and scene encoders produce two visual representations:
\begin{equation}
\mathbf{h}_b = f_b(x_b), \qquad \mathbf{h}_s = f_s(x_s),
\end{equation}
where $f_b(\cdot)$ is the ResNet-18 body encoder and $f_s(\cdot)$ is the CLIP ViT-B/16 scene encoder. Both features are projected to the same dimension:
\begin{equation}
\mathbf{z}_b = W_b \mathbf{h}_b + \mathbf{b}_b, \qquad
\mathbf{z}_s = W_s \mathbf{h}_s + \mathbf{b}_s.
\end{equation}
The fusion block takes the projected body and scene features after concatenation. A small fully connected module learns the shared representation used by both prediction heads. This is feature-level fusion. It combines global stream features and does not perform patch-level cross-attention between body and scene regions. The fused feature is computed as:
\begin{equation}
\mathbf{z} = \phi([\mathbf{z}_b;\mathbf{z}_s]),
\end{equation}
where $[\cdot;\cdot]$ denotes concatenation and $\phi(\cdot)$ is the fusion block. The categorical and VAD heads are defined as:
\begin{equation}
\hat{\mathbf{y}} = \sigma(W_c\mathbf{z}+\mathbf{b}_c), \qquad
\hat{\mathbf{v}} = W_v\mathbf{z}+\mathbf{b}_v,
\end{equation}
where $\hat{\mathbf{y}}\in[0,1]^{26}$ gives the predicted emotion probabilities and $\hat{\mathbf{v}}\in[0,1]^3$ gives the predicted valence, arousal, and dominance values.

\begin{figure*}[!t]
\centering
\includegraphics[width=0.94\textwidth]{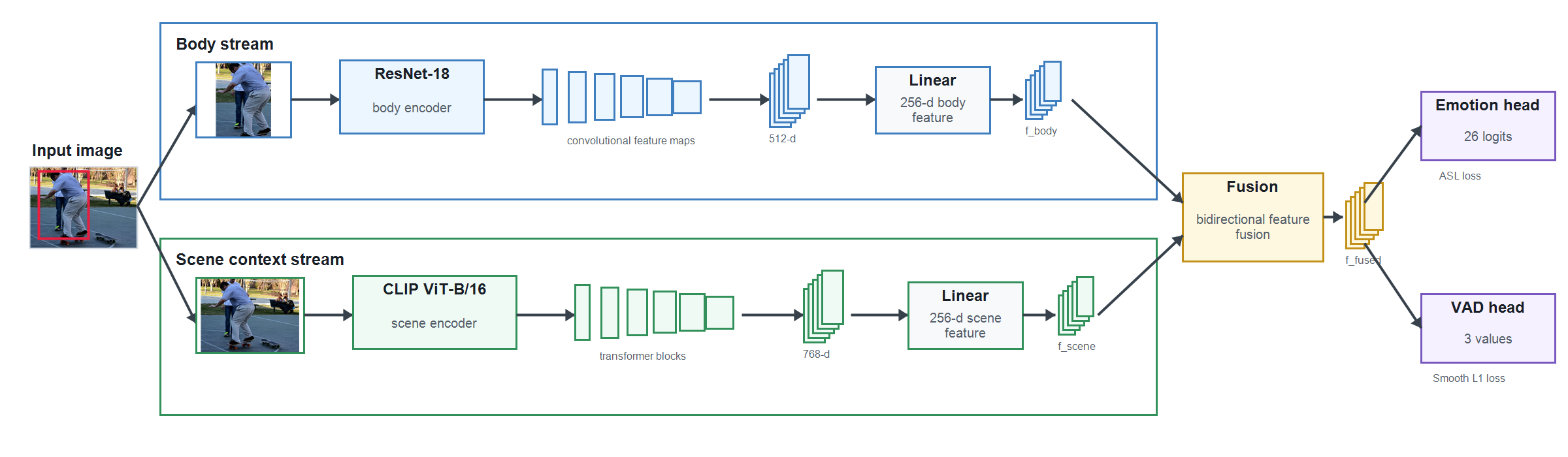}
\caption{Final two-stream architecture. The body branch encodes the target-person crop with ResNet-18, while the context branch encodes the full image with CLIP ViT-B/16. The projected features are fused and passed to separate categorical emotion and VAD regression heads.}
\label{fig:architecture}
\end{figure*}

Fig.~\ref{fig:architecture} shows the full architecture. The separation is simple: ResNet-18 focuses on the annotated person, and CLIP ViT-B/16 reads the whole scene.

\subsection{Loss Function}
The model is trained with a joint loss:
\begin{equation}
\mathcal{L} = \mathcal{L}_{cat} + \mathcal{L}_{vad}.
\end{equation}
Both losses are weighted equally in the current implementation. The categorical labels use Asymmetric Loss (ASL), which is suitable for long-tailed multi-label classification \cite{ridnik2021asl}. VAD uses Smooth L1, also known as Huber loss \cite{huber1964robust}. Optimization uses AdamW \cite{loshchilov2019adamw}, warmup, cosine decay, gradient clipping, MixUp \cite{zhang2018mixup}, dropout regularization \cite{srivastava2014dropout}, and early stopping based on validation mAP.

For the multi-label categorical task, each sample has a binary label vector $\mathbf{y}\in\{0,1\}^{26}$. In simplified form, ASL applies different focusing weights to positive and negative labels:
\begin{equation}
\mathcal{L}_{cat}
= - \sum_{k=1}^{26}
y_k(1-\hat{y}_k)^{\gamma_+}\log(\hat{y}_k)
 + (1-y_k)\hat{y}_k^{\gamma_-}\log(1-\hat{y}_k),
\end{equation}
where $\gamma_+$ and $\gamma_-$ control the positive and negative focusing terms. For VAD regression, Smooth L1 loss is used over the three continuous dimensions:
\begin{equation}
\mathcal{L}_{vad}
= \frac{1}{3}\sum_{d=1}^{3}\mathrm{SmoothL1}(\hat{v}_d-v_d).
\end{equation}

\subsection{Ablated Extensions}
Several additions are tested around the two-stream baseline. One variant adds a simplified CCIM-style context intervention module. Another, CLEF-lite, subtracts a learned context-bias output from the full-scene feature path. Rare-class training changes are also tested through ASL tuning and class-balanced sampling, following the general class-imbalance motivation used in class-balanced learning \cite{cui2019classbalanced}. These are lightweight ablations inspired by CCIM and CLEF, not exact reproductions, because the original methods include specific causal or counterfactual branch designs and training procedures.

\subsection{Training Setup}
Table~\ref{tab:hyperparams} lists the main training settings. The encoders are frozen for the first five epochs. After that, the body encoder and the top CLIP blocks are fine-tuned with a smaller learning rate, and the best checkpoint is chosen by validation mAP.

\begin{table}[!htbp]
\centering
\caption{Main settings for the selected two-stream model.}
\label{tab:hyperparams}
\begin{tabular}{lr}
\toprule
Setting & Value \\
\midrule
Body encoder & ResNet-18 \\
Context encoder & CLIP ViT-B/16 \\
Feature dimension & 256 \\
Batch size & 16 \\
Epoch budget & 40 \\
Best checkpoint epoch & 21 \\
Freeze epochs & 5 \\
Head learning rate & $1\times10^{-4}$ \\
Backbone learning rate & $1\times10^{-5}$ \\
Dropout & 0.55 \\
ASL $\gamma_{neg}$ & 4.0 \\
MixUp $\alpha$ & 0.3 \\
\bottomrule
\end{tabular}
\end{table}

\subsection{Evaluation Metrics}
For categorical emotions, the evaluation uses mean Average Precision (mAP), the main EMOTIC metric for multi-label classification. Hamming loss is reported with a fixed threshold of 0.5. The validation threshold that maximized F1 is also reported. For VAD regression, the metrics are Mean Absolute Error (MAE) and Root Mean Squared Error (RMSE).

The categorical score is computed as the mean of the Average Precision values over all emotion classes:
\begin{equation}
\mathrm{mAP} = \frac{1}{26}\sum_{k=1}^{26}\mathrm{AP}_k.
\end{equation}
For VAD regression, MAE and RMSE are computed over all test samples and the three affective dimensions:
\begin{equation}
\mathrm{MAE} = \frac{1}{3N}\sum_{i=1}^{N}\sum_{d=1}^{3}
|\hat{v}_{i,d}-v_{i,d}|,
\end{equation}
\begin{equation}
\mathrm{RMSE} =
\sqrt{\frac{1}{3N}\sum_{i=1}^{N}\sum_{d=1}^{3}
(\hat{v}_{i,d}-v_{i,d})^2}.
\end{equation}

\section{Results}
The final test result is reported first, followed by class-level behavior, ablations, and comparison with earlier EMOTIC work. The numbers answer the three research questions directly. The clean CLIP-based two-stream model is the reference point for all later comparisons.

\subsection{Overall Test Performance}
The clean two-stream model has the highest score among the tested variants. On the EMOTIC test split, it reaches 34.52\% mAP. No tested extension improves over this score. Table~\ref{tab:main_results} gives the main metrics.

\begin{figure}[!htbp]
\centering
\includegraphics[width=0.96\linewidth]{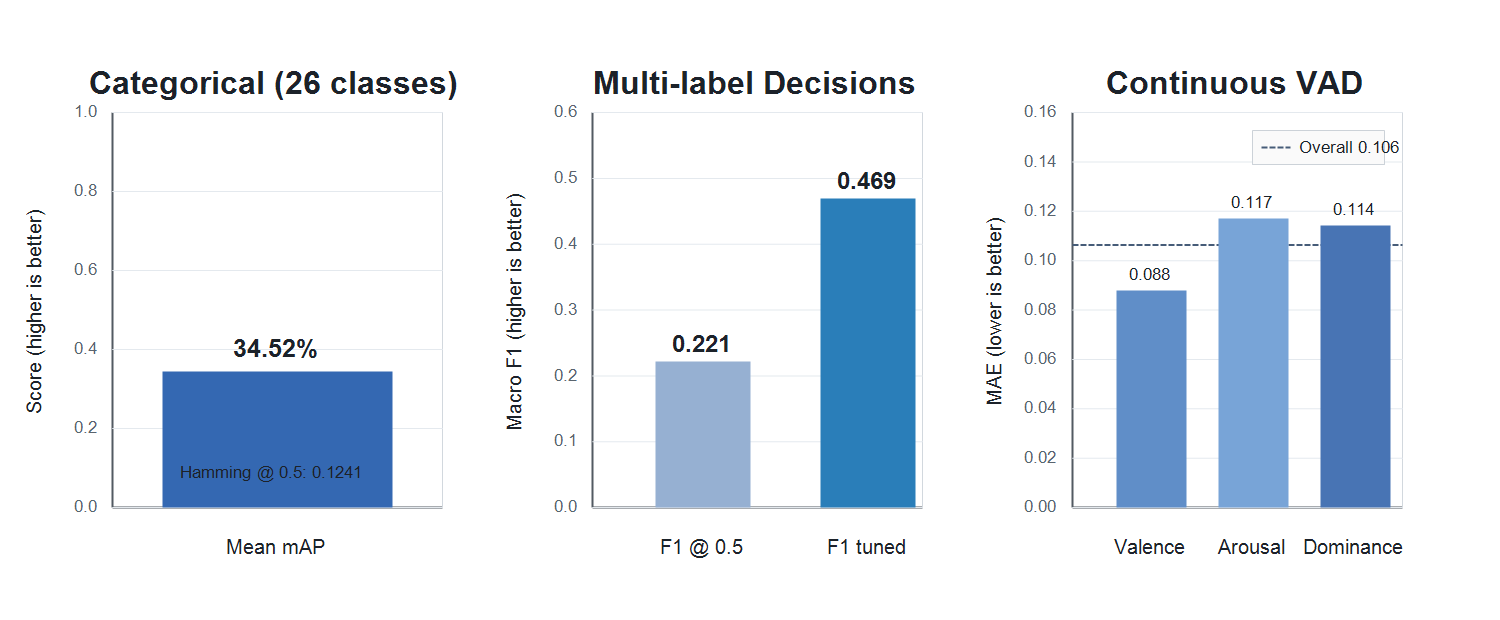}
\caption{Metrics dashboard for the final two-stream checkpoint.}
\label{fig:metrics}
\end{figure}

\begin{figure*}[!t]
\centering
\includegraphics[width=0.84\textwidth]{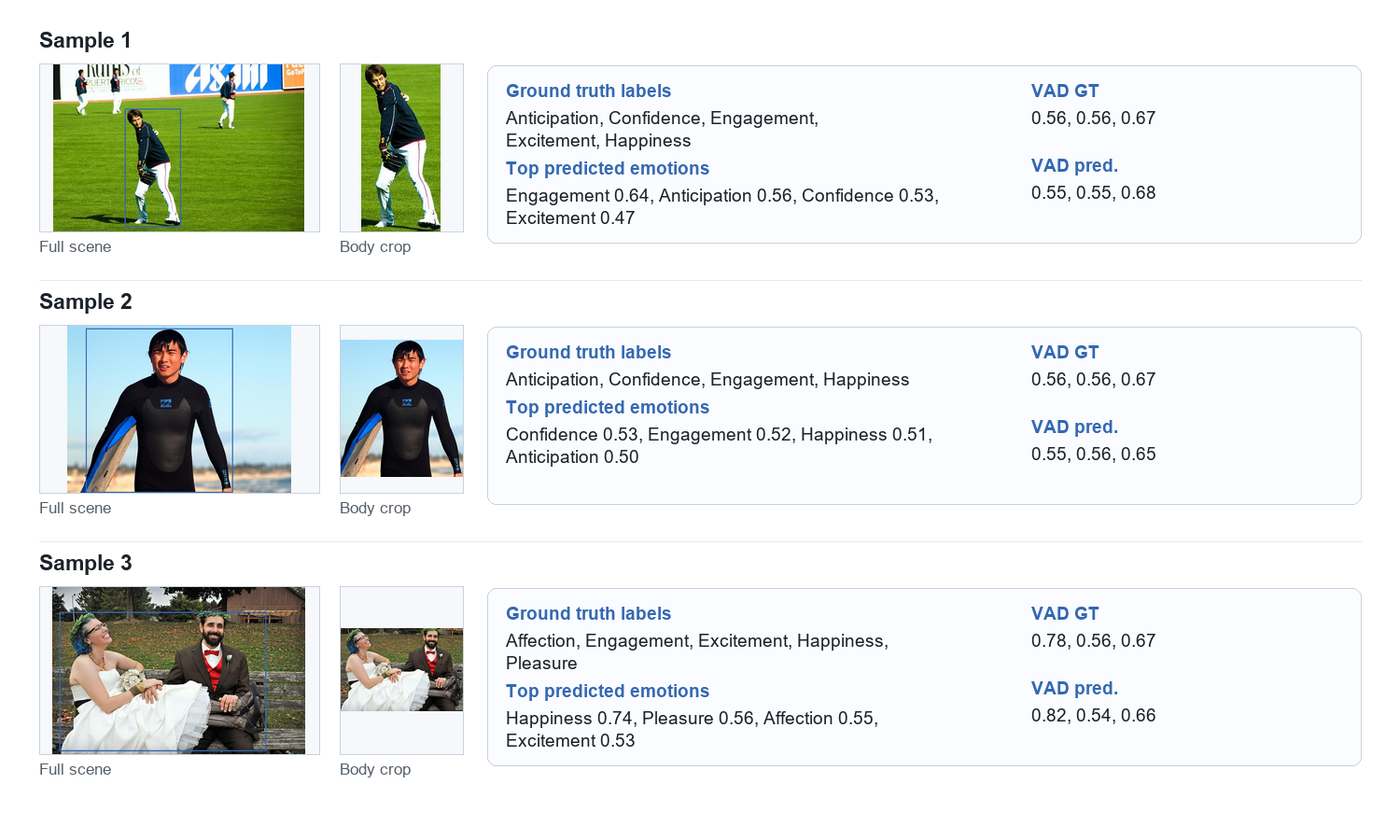}
\caption{Representative qualitative test predictions from the final two-stream model. Each example shows the full image with the target-person box, the body crop used by the body stream, ground-truth labels, top predicted emotion probabilities, and normalized ground-truth/predicted VAD values.}
\label{fig:sample_predictions}
\end{figure*}

Fig.~\ref{fig:metrics} shows the same result as a metrics dashboard, including categorical mAP, Hamming loss, and VAD regression error. Fig.~\ref{fig:sample_predictions} adds qualitative examples with the full scene, target body crop, ground-truth labels, top predicted emotions, and normalized VAD values. These are representative examples, not a complete qualitative evaluation, and some cases do not place every ground-truth label among the top displayed predictions.

\begin{table}[!htbp]
\centering
\caption{Test performance of the selected model.}
\label{tab:main_results}
\begin{tabular}{lr}
\toprule
Metric & Result \\
\midrule
Categorical mAP & 34.52\% \\
Hamming loss at 0.5 & 0.1241 \\
VAD MAE & 0.1064 \\
VAD RMSE & 0.1365 \\
Best validation threshold & 0.3526 \\
Best validation F1 & 0.4695 \\
\bottomrule
\end{tabular}
\end{table}

\subsection{Per-Class Behavior}
The class-level results are uneven, which is expected on EMOTIC. The model does well on common or visually supported categories such as Engagement, Happiness, Confidence, Excitement, and Anticipation. It is much weaker on rare or subtle categories, including Embarrassment, Yearning, Surprise, Fear, and Aversion. Global body-scene features are enough for broad affective cues. They are not enough for every fine-grained label.

\begin{figure*}[!t]
\centering
\includegraphics[width=0.70\textwidth]{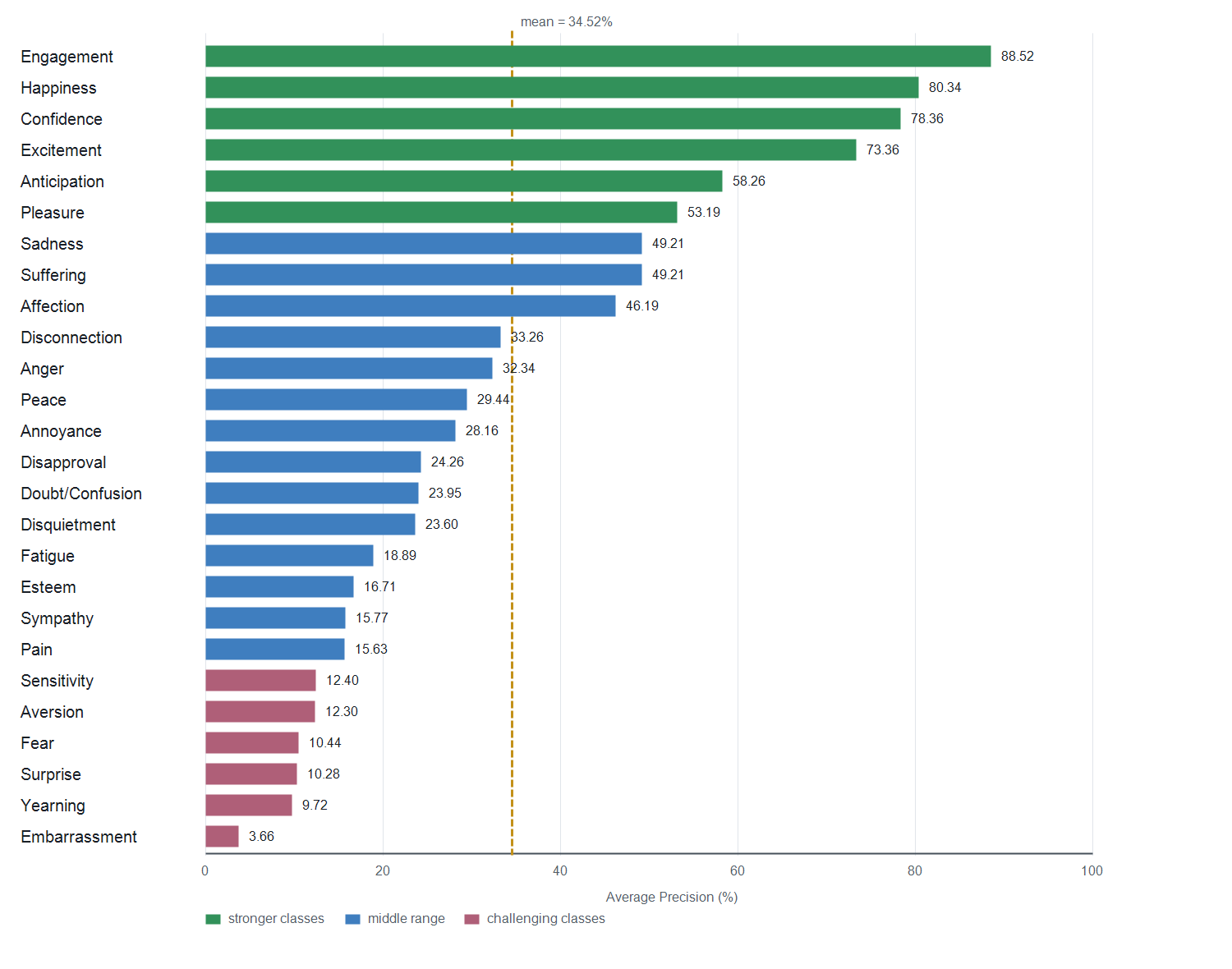}
\caption{Per-class Average Precision (AP) for all 26 EMOTIC emotion categories from the final two-stream checkpoint.}
\label{fig:per_class}
\end{figure*}

Table~\ref{tab:per_class} lists representative strong and challenging classes. Fig.~\ref{fig:per_class} gives the full AP distribution over all 26 emotion categories.

\begin{table}[!htbp]
\centering
\caption{Selected per-class AP values for the final model.}
\label{tab:per_class}
\begin{tabular}{lrlr}
\toprule
Strong class & AP & Challenging class & AP \\
\midrule
Engagement & 88.52 & Embarrassment & 3.66 \\
Happiness & 80.34 & Yearning & 9.72 \\
Confidence & 78.36 & Surprise & 10.28 \\
Excitement & 73.36 & Fear & 10.44 \\
Anticipation & 58.26 & Aversion & 12.30 \\
Pleasure & 53.19 & Sensitivity & 12.40 \\
\bottomrule
\end{tabular}
\end{table}

\subsection{Ablation Study}
Table~\ref{tab:ablations} gives the main two-stream ablations. The clean model reaches 34.52\% mAP. The simplified CCIM-style module drops to 34.04\% mAP. CLEF-lite, ASL tuning, and balanced sampling are closer, but still below the clean baseline.

\begin{table*}[!t]
\centering
\caption{Two-stream ablation results on the EMOTIC test split.}
\label{tab:ablations}
\renewcommand{\arraystretch}{1.18}
\begin{tabular}{p{3.4cm}p{7.3cm}cc}
\toprule
Experiment & What changed & mAP & Reading \\
\midrule
Two-stream & Clean ResNet-18 body + CLIP scene baseline & \textbf{34.52} & Highest tested score \\
Two-stream + simplified CCIM & Added a simplified context intervention module to the CLIP scene feature & 34.04 & Did not improve this setup \\
ASL tuning + balanced sampling & Adjusted ASL and used class-balanced sampling & 34.46 & Near-neutral \\
CLEF-lite & Subtracted a learned context-bias output from full-scene features & 34.42 & Near-neutral \\
\bottomrule
\end{tabular}
\end{table*}

These ablations changed the interpretation of the project. Context-debiasing modules were expected to help, because recent papers report gains from that direction. In this setup, the CLIP scene feature was already doing a lot of work. The simplified CCIM-style module did not improve the CLIP-based baseline, and CLEF-lite, ASL tuning, and balanced sampling stayed near the same range without passing it. Without multiple seeds, the small differences among the close variants should be read cautiously.

\subsection{Comparison with Prior Work}
Table~\ref{tab:prior_work} places this result beside related EMOTIC work. This is context, not a strict leaderboard. The compared papers differ in backbone, modalities, supervision, training details, and evaluation implementation. Some recent methods also report relative gains over their own base models rather than one directly matched CLIP-based two-stream setting.

\begin{table*}[!t]
\centering
\caption{Contextual comparison with related EMOTIC work. Reported values are not a strict leaderboard because the compared methods use different backbones, modalities, and evaluation settings.}
\label{tab:prior_work}
\renewcommand{\arraystretch}{1.18}
\begin{tabular}{p{3.1cm}p{7.3cm}c}
\toprule
Method & Main idea & Reported EMOTIC AP/mAP \\
\midrule
Kosti et al. \cite{kosti2017emotion,kosti2020context} & Person crop and scene CNN fusion for emotion recognition in context & $\sim$27--28 \\
EmotiCon \cite{mittal2020emoticon} & Multimodal context with personal, semantic, and socio-dynamic cues & 35.48 \\
CCIM \cite{yang2023contextdeconfounded} & Causal intervention to reduce context confounding & Relative gains over several base models \\
CLEF \cite{yang2024clef} & Counterfactual removal of direct context bias & Relative gains over several base models \\
DSCT \cite{li2024dsct} & Single-stage subject-context transformer with decoupled queries & 6.46 AP gain over its two-stage baseline \\
Ours & Image-only ResNet-18 body stream + CLIP scene stream & 34.52 \\
\bottomrule
\end{tabular}
\end{table*}

For CCIM and CLEF, the table uses qualitative wording because their results are presented as improvements over multiple base models rather than a single directly matched CLIP-based two-stream setting. For DSCT, the entry is the relative gain stated by the paper, not a directly comparable absolute mAP for this setting.

\section{Discussion}
The CLIP-based scene branch gives the model a good global reading of the scene. The debiasing-style variants were expected to improve on top of that, but they did not. One likely reason is architectural. CLIP already produces a global, semantically dense full-scene representation, so a context-bias subtraction or intervention module may be trying to remove something that overlaps with the signal needed for prediction. In that case, the correction becomes redundant, or worse, partly destructive. This does not prove that causal or counterfactual debiasing is ineffective in general. The claim is narrower: simplified adaptations may not be enough when they are placed on top of a CLIP-based scene representation that already carries a lot of semantic information.

The comparison with heavier context-aware systems shows the trade-off. EmotiCon uses multiple context streams for personal, semantic, and socio-dynamic information. The proposed model uses one CLIP-based scene branch and one ResNet-18 body stream. That simpler design does not replace explicit depth, object-relation, or interaction modeling, but it reaches a competitive range among earlier context-aware approaches while remaining easier to implement. CLIP can cover part of the semantic context work. It cannot do all the structured reasoning.

The remaining errors make that limit visible. Embarrassment, Yearning, Surprise, Fear, and Aversion are not only rare; they are also hard to infer from global appearance. They may depend on fine-grained social cues, the object a person is using, another person's reaction, or subjective interpretation. Adding another global branch is unlikely to solve that by itself. Label-correlation modeling and fine-grained subject-context interaction are more plausible next steps.

\section{Limitations}
This work has several limitations. The model predicts apparent emotion from images, not a person's true internal emotional state. It also assumes target-person bounding boxes and does not solve subject detection. The CLIP scene stream encodes the full image globally, so irrelevant background can enter the representation, and the model does not explicitly identify which object, person, or scene region matters for the target subject. The CCIM-style and CLEF-lite variants are simplified adaptations rather than exact reproductions of the original causal and counterfactual methods. Finally, all experiments are on EMOTIC, and small differences between close mAP values should be read cautiously without repeated runs.

\section{Future Work}
The next experiment should start from the clean CLIP-based two-stream model, not from the simplified debiasing variants. The most direct addition is a lightweight label-correlation head. EMOTIC is multi-label, and emotions such as Happiness, Excitement, Pleasure, and Engagement often co-occur. A head that models these relationships would target the label structure directly instead of asking the image encoder to learn everything alone.

Fine-grained subject-context interaction is the second direction. The model should learn which scene regions, objects, or nearby people are emotionally relevant to the target person. The current fusion block cannot do that. It sees a body vector and a scene vector.

The body stream can also be strengthened. The current model uses ResNet-18 with $128\times128$ body crops; future experiments can test ResNet-50 or EfficientNet backbones and increase the body input size to $224\times224$ to preserve more posture and clothing detail. Longer-term work can explore patch-level body-scene attention, object relation modeling, and a DSCT-style single-stage model.

\begin{figure}[!htbp]
\centering
\includegraphics[width=0.92\linewidth]{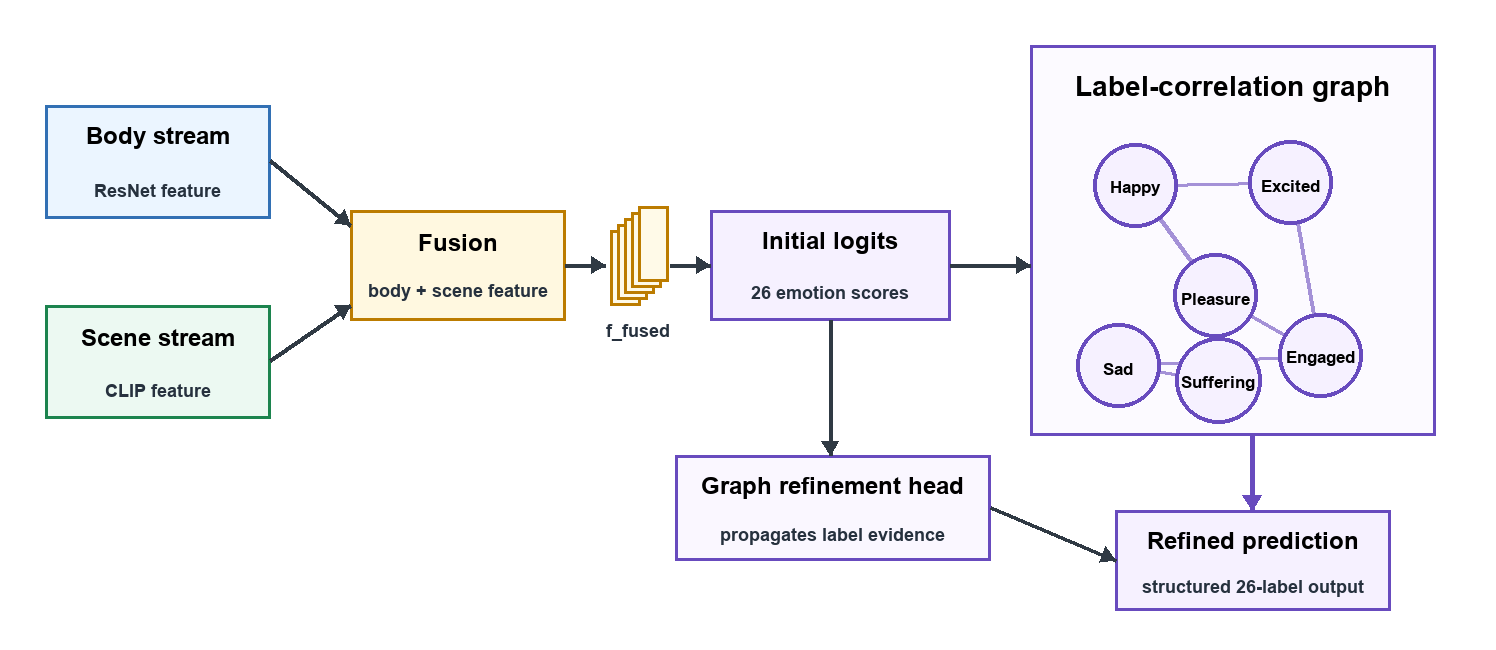}
\caption{Proposed future extension: a lightweight label-correlation graph head that can be added on top of the current two-stream CLIP baseline.}
\label{fig:future}
\end{figure}

Fig.~\ref{fig:future} sketches the label-correlation direction. The goal is to keep the two-stream visual backbone and add a small graph-based refinement step over emotion labels.

\section{Conclusion}
This work studied context-aware emotion recognition on EMOTIC with a two-stream body-scene model. The final system combines a ResNet-18 body stream with a CLIP ViT-B/16 full-scene stream and achieves 34.52\% mAP on the EMOTIC test split. None of the tested variants, including simplified CCIM-style intervention, CLEF-lite context-bias subtraction, and rare-class training changes, improves over the clean two-stream model. CLIP gives broad global scene semantics, but simplified causal, counterfactual, or sampling-based modifications do not automatically improve this baseline. The hard classes remain the rare and subtle ones. The next step is structure: label-correlation modeling and fine-grained subject-context interaction.

\bmhead{Statements and Declarations}

\textbf{Funding} The authors received no specific funding for this work.

\textbf{Competing interests} The authors declare that they have no competing interests.

\textbf{Data availability} The EMOTIC dataset used in this study is available from the original dataset providers subject to their access conditions. Additional result files generated during this study are available from the corresponding author on reasonable request.

\textbf{Code availability} The implementation used for the experiments is available from the corresponding author on reasonable request.

\textbf{Author contributions} Zubair Abbas designed and implemented the experiments, analyzed the results, and wrote the manuscript. Muhammad Umair supervised the study. Muqaddas Hameed conducted the literature review for the project. All authors read and approved the final manuscript.

\begingroup
\small
\sloppy
\bibliography{references}
\endgroup

\end{document}